\documentclass[letterpaper]{article} % DO NOT CHANGE THIS
\usepackage{aaai23}  % DO NOT CHANGE THIS
\usepackage{times}  % DO NOT CHANGE THIS
\usepackage{helvet}  % DO NOT CHANGE THIS
\usepackage{courier}  % DO NOT CHANGE THIS
\usepackage[hyphens]{url}  % DO NOT CHANGE THIS
\usepackage{graphicx} % DO NOT CHANGE THIS
\usepackage{natbib}  % DO NOT CHANGE THIS AND DO NOT ADD ANY OPTIONS TO IT
\usepackage{caption} % DO NOT CHANGE THIS AND DO NOT ADD ANY OPTIONS TO IT
\usepackage{algorithm}
\usepackage{adjustbox}
\usepackage{amsfonts}
\usepackage{newtxtext}
\usepackage{multirow}
\usepackage{multicol}
\usepackage{arydshln}
\usepackage{color, colortbl}
\usepackage{subcaption}
\usepackage{ctable}
\usepackage{pdfpages}
\definecolor{lightgray}{gray}{0.9}
\usepackage{placeins}
\usepackage{amsmath}
\usepackage{bbm}
\usepackage{algpseudocode}
\definecolor{cvprblue}{rgb}{0.21,0.49,0.74}
\usepackage[pagebackref,breaklinks,colorlinks,allcolors=cvprblue]{hyperref}
\usepackage{newfloat}
\usepackage{listings}
\DeclareCaptionStyle{ruled}{labelfont=normalfont,labelsep=colon,strut=off} % DO NOT CHANGE THIS
\floatstyle{ruled}
\newfloat{listing}{tb}{lst}{}
\floatname{listing}{Listing}
\title{Towards Better Visualizing the Decision Basis of Networks via \\Unfold and Conquer Attribution Guidance}
\author{
Jung-Ho Hong$^{1}$\textsuperscript{\rm \dag},
Woo-Jeoung Nam$^{2}$\textsuperscript{\rm \dag},
Kyu-Sung Jeon$^{1}$,
Seong-Whan Lee$^{1}$\textsuperscript{\rm \footnote{Corresponding Author, $^\dag$ Equal Contribution}}
}
\affiliations{
$^{1}$Department of Artificial Intelligence, Korea University, Seoul, Republic of Korea\\
$^{2}$School of Computer Science and Engineering, Kyungpook National University, Daegu, Republic of Korea\\
\{jungho-hong, ksjeon, sw.lee\}@korea.ac.kr, nwj0612@knu.ac.kr \\ 
{\hypersetup{urlcolor=blue}
\fontsize{9pt}{12pt}\selectfont \href{https://github.com/KU-HJH/UCAG}{https://github.com/KU-HJH/UCAG}}
}

\usepackage{bibentry}
\begin{document}

\maketitle

\begin{abstract}
\begin{quote}
Revealing the transparency of Deep Neural Networks (DNNs) has been widely studied to describe the decision mechanisms of network inner structures.
In this paper, we propose a novel post-hoc framework, Unfold and Conquer Attribution Guidance (UCAG), which enhances the explainability of the network decision by spatially scrutinizing the input features with respect to the model confidence. 
Addressing the phenomenon of missing detailed descriptions, UCAG sequentially complies with the confidence of slices of the image, leading to providing an abundant and clear interpretation.
Therefore, it is possible to enhance the representation ability of explanation by preserving the detailed descriptions of assistant input features, which are commonly overwhelmed by the main meaningful regions. We conduct numerous evaluations to validate the performance in several metrics: i) deletion and insertion, ii) (energy-based) pointing games, and iii) positive and negative density maps. Experimental results, including qualitative comparisons, demonstrate that our method outperforms the existing methods with the nature of clear and detailed explanations and applicability.
\end{quote}
\end{abstract}

\section{Introduction}
% Deep learning의 성공
%% Interpretability의 중요성
Although Deep Neural Networks (DNNs) show remarkable performance in various computer science tasks due to their applicability, the issues of complexity and transparency derived from the non-linearity of the network hinder the clear interpretation of the decision basis. 
Visual explanations of network decisions alleviate these problems and provide analytical guidance for model development with intuitive explanations.
Inspired by this, several approaches have been introduced, such as relevance or gradient-based generation, to describe the image regions that the network currently references to make decisions.

\begin{figure}[h!]
\centering
\includegraphics[width=0.47\textwidth]{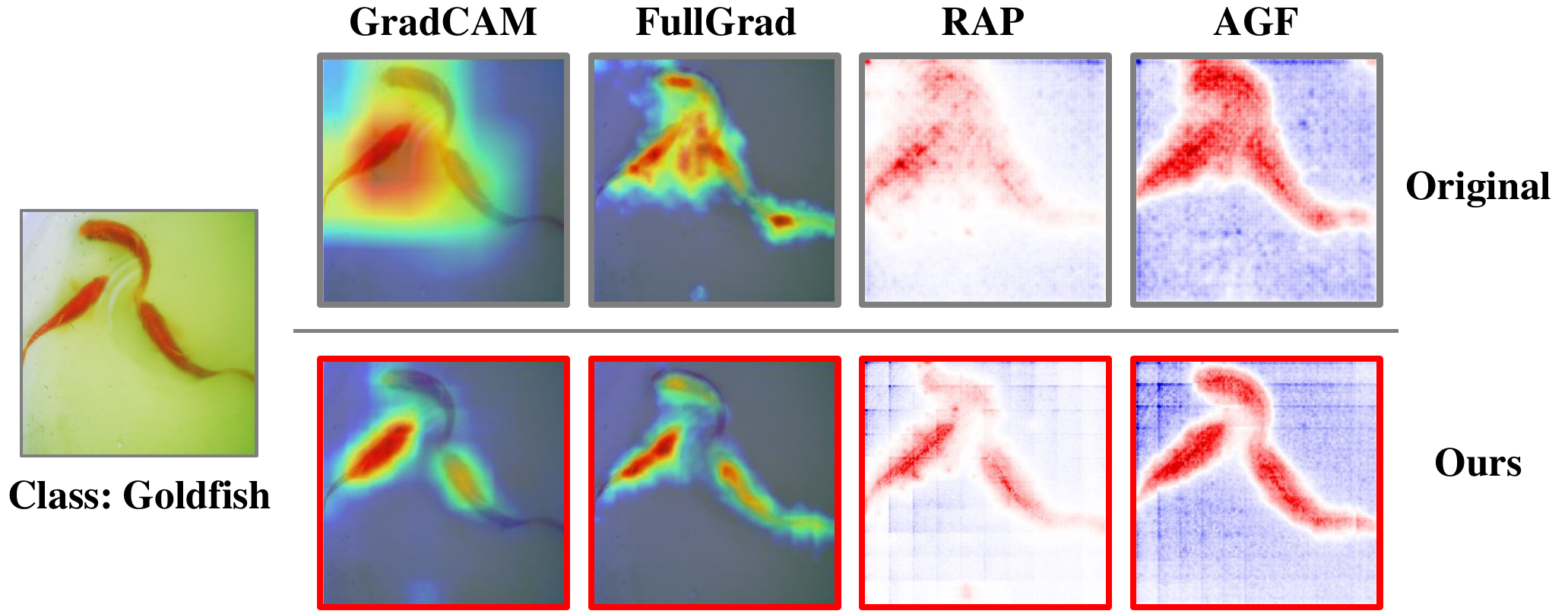}
\caption{The qualitative results of our approach. The first and second rows represent the existing visualization methods and ours, respectively. Our approach enhances the representation ability of explanation maps, which is applicable to any explanation technique.}
\label{fig:img1}
\end{figure}

The approaches behind the relevance score generate the explanation based on the decomposition methods, Layer-wise Relevance Propagation (LRP)~\cite{bach2015pixel}. LRP constructs the propagation rule according to the predicted scores of the specific class. Despite the nature of employing relevance score is robust to the shattered gradient problem and success in producing the input-level explanation map, the class-discriminativeness within their representation is still limited. Recently, although Attribution Guided Factorization (AGF) enhances the class discriminability of the explanation by integrating both gradient/relevance-based methodologies, it still provides biased attribution toward the class.

GradCAM~\cite{selvaraju2017grad} computes the coefficients for each map which are averaged over the gradients of the activation maps, and generates a class activation map by linearly combining the maps with the corresponding coefficients to provide a visual explanation. Although the explanation map generated from GradCAM successfully provides the class-discriminative explanation, the provided explanations are still coarse and contain numerous non-saliency regions due to the mechanism of utilizing the source map from the deepest layer, which keeps the lowest resolution activation map.

The compressive nature of DNNs limits the representation ability of relevance/gradient-based approaches. As the features in the deep layer of the network hold the abbreviated information, it restricts the propagated regions to represent abundant saliencies. Thus, the resulting saliency maps produced from the gradient or relevance approach are represented with limited quality, with obscured class-discriminative regions.
In this paper, we propose Unfold-and-Conquer Attribution Guidance (UCAG) framework to guide the explaining methods to generate better saliencies. It improves interpretability by investigating the detailed spatial regions with a set of subregions. Figure~\ref{fig:img1} illustrates the intuitive examples of our method, representing the strong advantages of detailed and abundant descriptions agnostic to models and methods.
To fulfill this, UCAG is divided into three sections: spatial unfoldment, partial saliency generation, and conquer with geometrical aggregation.
First, the unfoldment procedure provides a series of patches delivered by deploying the original data from a spatial perspective.
The attribute method independently computes the partial explanation map. During this procedure, UCAG accepts any specific type of methodology as an explanation module, including explaining mechanisms and model types.
% (애매)
Finally, the conquering procedure aggregates the partial saliency maps to produce the final explanatory map.
Therefore, UCAG encourages an explanation method that scrutinizes an image through the spatial axis.

The key contributions of this paper are summarized as:
\begin{itemize}
    \item We propose a new post-hoc framework: Unfold and Conquer Attribution Guidance (UCAG), to improve the interpretability of explanation methods by spatially scrutinizing the input data. Spatial scrutinization and integration provide multiple partial semantic views, leading to exploring the obscured saliencies due to the compressing nature of DNNs.
    
    \item Post-hoc manner of UCAG has the advantage of being easily applicable to various relevance/gradient-based approaches and DNN models. We present the applicability of UCAG that successfully provides a function of post-hoc refinement with intuitive and detailed descriptions of explanation.
    
    \item We validate the quality of the saliency map in density, localization, and causality perspectives. We conduct the verified experiments: i) positive and negative map density~\cite{jalwana2021cameras}, ii) energy-based and original pointing game~\cite{wang2020score}, and iii) insertion and deletion~\cite{petsiuk2018rise} to confirm the perspectives, respectively. The results demonstrate that our method outperforms the existing explaining methods with a sizable gap in terms of indicating the better points in the input correlated with the output decision.
    
\end{itemize}

\section{Related Work}
Gradient-based approaches~\cite{simonyan2013deep, zeiler2014visualizing, springenberg2014striving, selvaraju2017grad, chattopadhay2018grad, srinivas2019full,rebuffi2019normgrad, fu2020axiom, rebuffi2020there, zhang2021novel} construct the saliency map derived from the gradients. In the manner of employing class activation maps, which employs the activation map to generate the saliency map,
GradCAM~\cite{selvaraju2017grad} alleviates the limitations of CAM that require a global average pooling and re-training process.
They construct an explanatory map by linearly combining the activation map and the coefficients with the coefficients computed by taking the average of the gradients corresponding to the activation map.
XGradCAM~\cite{fu2020axiom} introduces the way to generate the class activation map while obeying two theoretical axioms: sensitivity and conservation.
WGradCAM~\cite{zhang2021novel} enhances the representation ability of the explanation map in terms of localization. 

Several approaches~\cite{fong2019understanding, jiang2021layercam, jalwana2021cameras, zhang2021novel} turned the literature to amplifying the quality of saliency maps to produce a more precise saliency map.
Extremal perturbation~\cite{fong2019understanding} generates the saliency map following the searching nature of perturbation-based approaches.
LayerCAM~\cite{jiang2021layercam} aggregates the information from earlier layers and introduces the way of deciding the coefficients of each source.
CAMERAS~\cite{jalwana2021cameras} fuses the multi-resolution images to generate a high-resolution explanation while obeying the model fidelity. 
SigCAM~\cite{zhang2021novel} combines perturbation/gradient-based approaches to produce more sparse and precise explanations using learnable external parameters.

Relevance-based approaches~\cite{bach2015pixel, zhang2018top, gu2018understanding, iwana2019explaining, nam2020relative, nam2021interpreting, gur2021visualization} produce the granular explanation by decomposing the output predictions in a backward manner. LRP~\cite{bach2015pixel} introduces the propagation rules that decompose the output logits into the relevance scores while maintaining the conservation rule. As the recent research, RAP~\cite{nam2020relative} aims to separate the relevant and irrelevant parts of input by allocating the bi-polar relevance scores during propagation. RSP~\cite{nam2021interpreting} extends the rule of RAP to overcome the limitations of suppressing nature, resulting in solving the class-specific issues. AGF~\cite{gur2021visualization} combines the accumulated gradients and relevance scores across the layers to correct the saliency bias.

Despite the superior performance of the previous approach, the generated explanation maps represent insufficient saliency and diminished class discrimination.
To increase the fidelity of the explanation maps, we consider the schemes of spatial scrutinization and integration by providing multiple partial information from a single input image, leading to discovering the obscured regions while maintaining class-discriminative saliencies.

% \section{The Proposed Methods}
% In this section, we describe the problems with the re-scaling procedure in the various models and CAM modules, then introduce our approach to relieve this issue. 
\section{Saliency Shedding}
Propagation-based approaches provide a saliency map with the same size as the input image by propagating the relevance score with respect to the network decision. However, it is difficult to clarify the correlations of neurons among the whole layers, resulting in inevitably assigning positive attributions to unrelated pixels. 
On another side, since the CAM-based approach provides the explanation based on the activation map of the deepest feature extraction layer and gradients, the computed saliency map represents a coarse explanation and lacks details due to the upsampling. To provide more detailed saliency maps, upscaling techniques of the input image are introduced in~\cite{jalwana2021cameras} to increase the resolution of the class activation map.

\begin{figure}[!tbp]
% \includepdf[width=1.0\linewidth]{./figs/figure-motiv.pdf}
\includegraphics[width=1.0\linewidth]{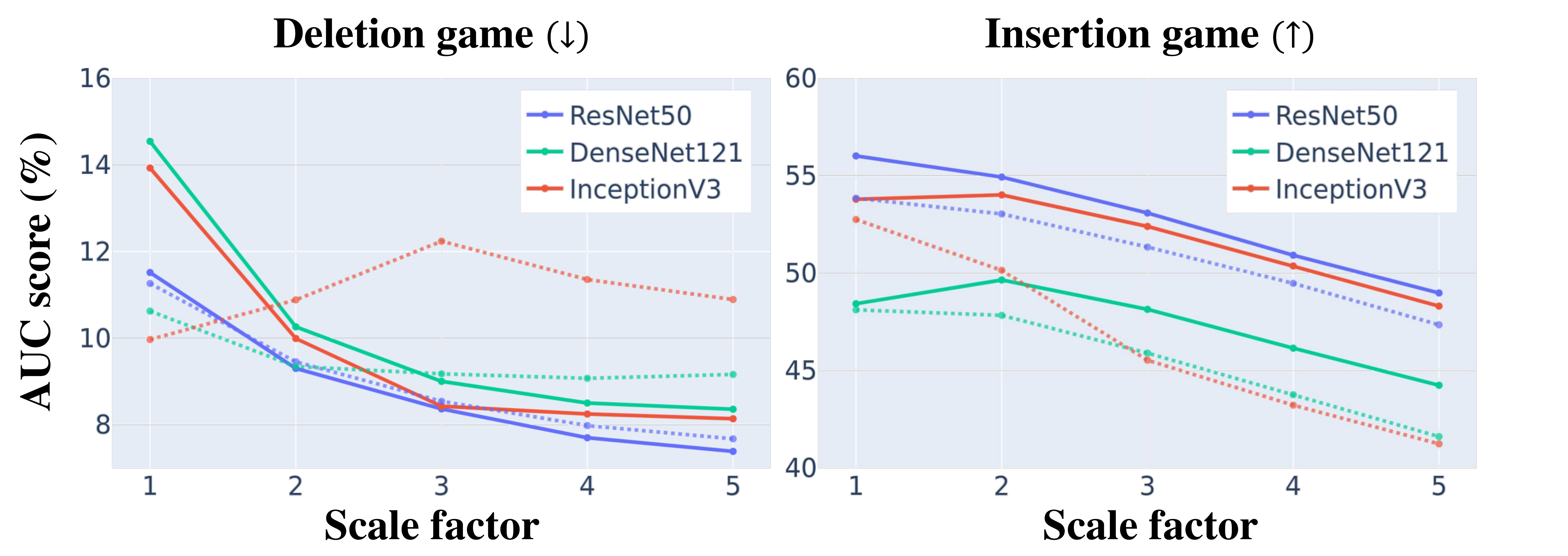}

\caption{The motivations of our work. The scores of deletion and insertion games among various resolutions with GradCAM on the ImageNet dataset. We scaled the size of image from $(224, 224)$ to $(1120, 1120)$ utilizing the scale factor $\alpha$. The performance of the original and our methods are represented as dots and solid lines, respectively. We aim to present a resolution-agnostic method with the enhancement of interpretability.}

\label{fig:ins_del}
\end{figure}

However, as shown in our empirical experiments of Figure~\ref{fig:ins_del}, employing the upsampled images both decreases the deletion and insertion scores proportional to the resolution, which are originally in the trade-off relationships. Deletion and insertion scores are widely used metrics in the field of explainability ~\cite{petsiuk2018rise,wang2020score,zhang2021novel,jung2021towards}, to evaluate whether the explanation map successfully covers the more fine-grained representation and localizes the actual object better.
Decreasing insertion scores according to the upscaling states that the visualized explanation loses saliency areas due to the distortion of neuron activation and corresponding gradients.

Motivated by this issue, we address the method to improve the quality of saliency maps by investigating the correlations among the input unfolding and confidence scores, leading to maintaining the salient regions while increasing the granularity of the explanation map.

\begin{figure*}[!th]
\centering
\includegraphics[width=1.0\textwidth]{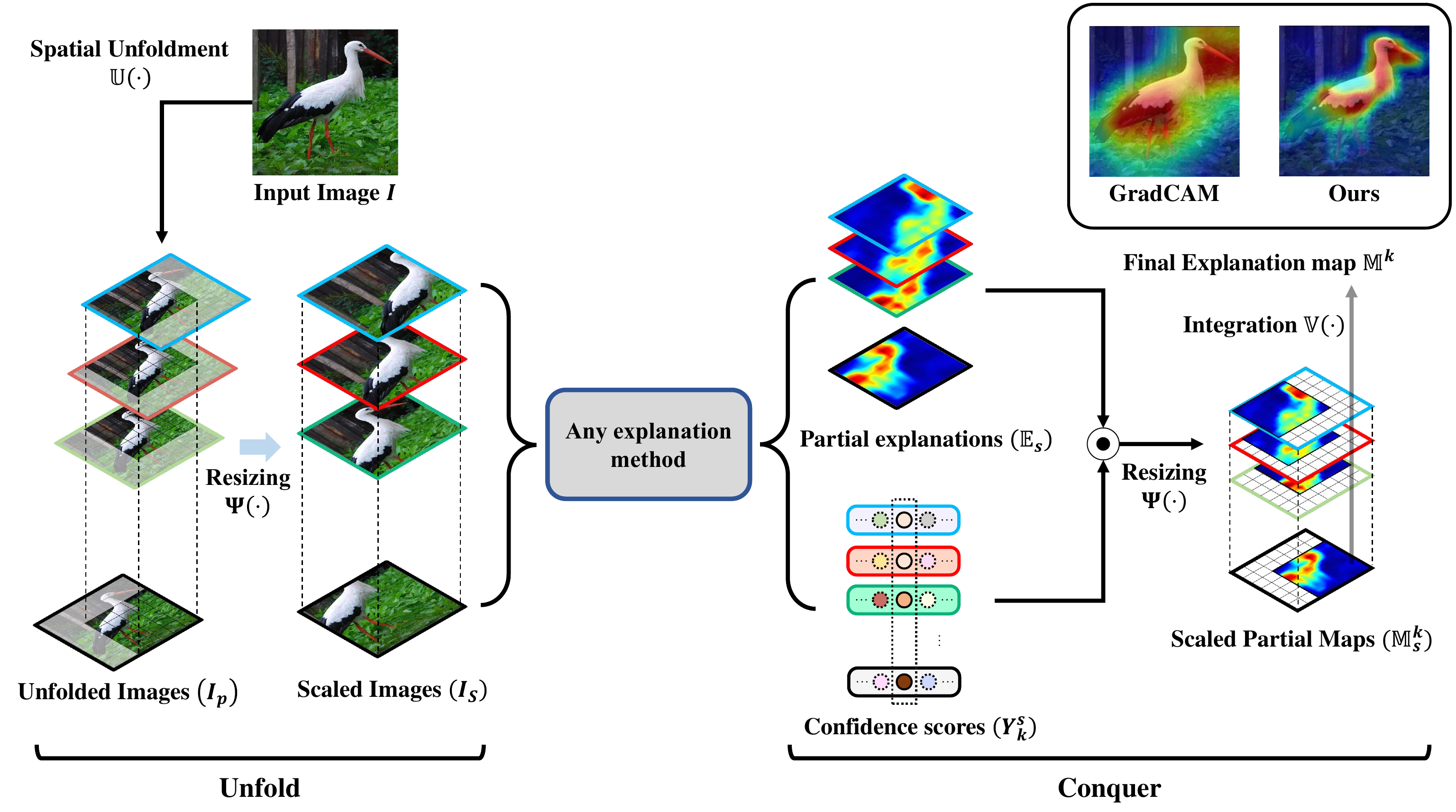}

\caption{An overview of UCAG. The entire course is divided into three streams: i) unfoldment, ii) patch-wise saliency generation, and iii) conquer with aggregation. Through the unfolding process, the unfolded patches are individually delivered to the gradient/propagation-based visualization methods. Conquering is a procedure for integrating segmented explanations into a single explanation map with the corresponding confidence scores.}

\label{fig:method}
\end{figure*}
\section{Unfold and Conquer Attribution Guidance}

To enhance the granularity of the explanation without losing saliencies, our work starts from the assumption that if the highlighted regions obviously contribute to the decision, their saliency should remain regardless of the geometrical translation. To fulfill this, we extend the aspect of the convolution process generally used in the DNNs framework.
Hence, we unfold the image into a series of segments and let the visualization method infers each fragment independently to yield partial explanation maps. By folding partial explanations weighted with corresponding confidence scores, the unified explanation contains the saliencies referring to the various perspectives of spatial views. Figure~\ref{fig:method} illustrates the overview of our method: UCAG. A detailed description of each stage is described in each section.
% Therefore, Our approach improves the quality of the saliency map through scrutinization by regarding the information provided from various spatial evidences.

% \subsubsection{Geometric Tessellation}
\subsection{Spatial Unfoldment}
% 용어 정의 및 과정
% To implement divide and conquer~
In this section, we demonstrate a strategy for providing the unfolded segments along the spatial axis to construct the set of input materials for the explanation method. To accomplish this, we first declare the hyper-parameter $n$: the number of segmented patches. Then, we construct an unfolding function: $\mathbb{U}$, which striding the input image $I\in \mathbb{R}^{C \times H \times W}$ into the series of $n^2$ patches $I_p$ by walking along the spatial (weight and height) axis. 
The set of patches includes the partial image $I_p^j \in \mathbb{R}^{C \times H' \times W'}$, indexed by $j \in \{0,1, \dots, n^2-1\}$. The size of patch $(H', W')$ is computed with the specific resizing factor $\rho \in [0, 1]$, described as $(\rho H, \rho W)$.

However, from the perspective of the network, utilizing the downscaled input leads to representing coarse output features after passing through the network feature extraction. To relieve this, we enhance the resolution of each patch with the scaling function $\Psi(I_p, \alpha)$, which scales the set of patches into the size of $(\alpha H', \alpha W')$.
In summary, the unfoldment procedure provides partial explanations as follows:
\begin{equation}
    % P^{k} = \mathcal{D}(I, \Theta, n). 
    I_s=\Psi(\mathbb{U}(I, \rho, n), \alpha).
    \label{eq:unfold}
\end{equation}

% Here, $\rho$ and $n$ denote the scaling value and the number of segments, respectively. The height and width of receptive size are determined as $H'$ and $W'$ is same as $\rho\times H$ and $\rho\times W$, respectively, working as the receptive semantic for the attribution methods.
Hence, the input image $I$ is unfolded into the set of patches $I_s$ with a size of $(n^2, C, \alpha{H^\prime}, \alpha{W^\prime})$ and each patch shares the overlapped areas with the striding mechanisms following the nature of unfoldment. From the spatial perspective, a set of patches holds different subregions from the single image and enables a network to review the various views.

\subsection{Patch-wise Saliency Generation}
In this section, we carefully revisit the general explaining methods: $\mathbb{E}$ with the two steams of gradient/attribution and clarify the applying process to our post-hoc frameworks.
\subsubsection{Gradient-based saliency}
To generate the saliency map revealing network decisions, gradient-based methods utilize the activation map in the defined layers and their corresponding neuron importance derived from the gradients. Generally, the activation map is obtained from the last layer of DNN before the linear or global average pooling layer to investigate the compressed feature importance. Several CAM approaches introduce their intrinsic way of computing the class-specific coefficient by utilizing the gradient of each pixel in the activations. Generic gradient-based saliency map~\cite{selvaraju2017grad} is computed as follows:

\begin{equation}
    \mathbb{E}_{grad}(I, f, k)=\textrm{ReLU}\left(\sum_c{w_c^k}A_c\right).
\end{equation}

Here, $c$, $f$, and $k$ denote the index of the channel axis, the kind of DNNs, and the index of the target class, respectively. $w_c^k$ is the neuron importance of each feature map derived from the gradients with respect to the target class. According to the provided assumption of existing approaches, e.g.~\cite{selvaraju2017grad,chattopadhay2018grad, fu2020axiom, zhang2021novel}, there is a slight difference in deriving the $w_c^k$, but mostly dependent on the chain rule ${\partial f(I)^{k}}/{\partial A}$. Detailed expansion of each method is described in supplementary materials.
In our post-hoc manners: using the unfolded segments $I_s$ as input, partial explanations are generated independently with the activations computed from each patch as $\mathbb{E}_{grad}(I_s, f, k)$.
% \begin{equation}
%     M^l_p=\textrm{ReLU}\left(\sum_c{w_c^{k,l}}A_c^l\right)
% \end{equation}
% Where the $l$ denotes the index of patch and $P$ denotes the activations computed with partial segments $I_s$. The partial activations $P$ and corresponding coefficients $w$ of each segment are computed independently using the input set of patches. The partial activations P and corresponding coefficients w of each segment are computed independently using the input set of patches. s

\subsubsection{Propagation-based saliency} 
Propagation-based methods are capable of providing the input-level explanation map, allowing for the representation of pixel-level attribution. They compute the relevance score by layer-wisely propagating the output logits to input in a backward manner. During the propagation, the whole relevance sum $R$ is maintained with the conservation rule~\cite{bach2015pixel}, acting as the evidence of the decision. The generic propagation rule~\cite{montavon2017explaining} is defined as follows:
\begin{equation}
\begin{split}
    % R_j^{(n)}&=\mathcal{G}_j^{(n)}\left(\bold{X},\Theta,R^{(n-1)}\right)\\
    % &=\sum_i\bold{X}_j\frac{\partical L^{(n)}_i(\bold{X},\Theta)}{\partical \bold{X}_j}\frac{R^{(n-1)}}{\sum_{i'}L^{(n)}_{i'}(\bold{X},\Theta)}
    \mathbb{E}_{att}(I, f, k) &= \mathcal{G}(I, \theta, R^{(l-1)})\\
    &= \sum_i A_j\frac{\partial f_i^{(l)}(A,\theta)}{\partial A_j}\frac{R_i^{(l-1)}}{\sum_{i^\prime}f_{i^\prime}^{l}(A,\theta)}.
\end{split}
\label{eq:relevance}
\end{equation}

Here, $l$ indexes the middle layer of DNNs, and $A$ represents the activation of corresponding layers. $i$ and $j$ denote the neuron index in the $l$ and $l-1$ layers, respectively. During the propagation, the whole relevance sum is maintained as $\sum_j R_j^{(l)} = \sum_i R_j^{(l-1)}$, starting from the output one-hot vector: $f(I)^k$ to the input layer.

Several kinds of propagation-based methods, e.g., SGLRP, c*LRP, RSP, and AGF are mainly based on Eq.~\ref{eq:relevance}, but have some techniques, such as contrastive perspective, uniform shifting, and factorization, to enhance the quality of attribution maps. Our method is independent of each of these unique techniques as $\mathbb{E}_{att}(I_s, f, k)$ and could be utilized for intuitive and clear visualizations.

\begin{figure*}[!ht]
\centering
\includegraphics[width=1.\textwidth]{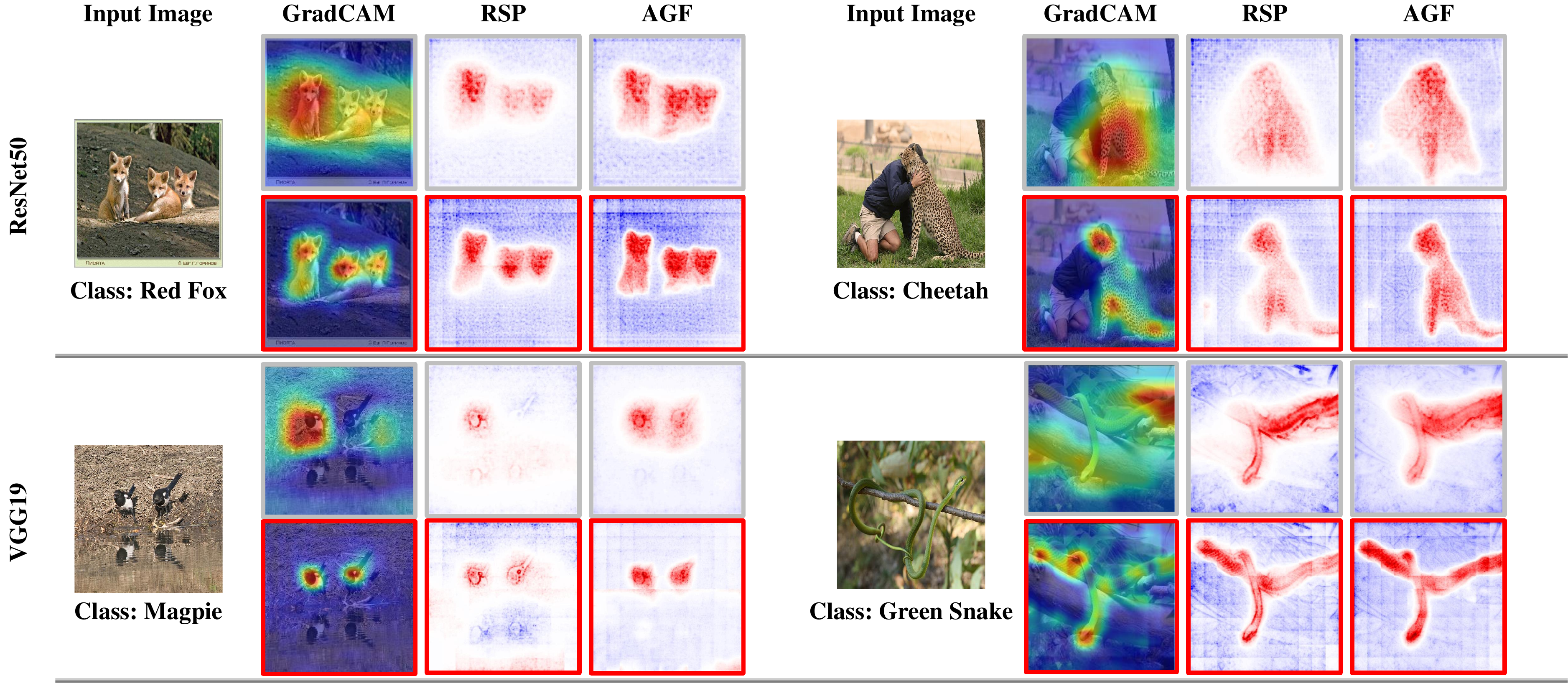}

\caption{Qualitative comparisons in VGG-19 and ResNet-50. UCAG improves the quality of the saliency map and has an advantage of agnostic to model and method. First and second rows in each group represent the original and our (marked as red) results, respectively. Additional results are illustrated in supplementary materials.}

\label{fig:result1}
\end{figure*}

\subsection{Conquer with Geometrical Aggregation}
% \subsection{Conquer to }

As the unfolded images do not guarantee to contain the actual object, it is necessary to discriminate the patches according to the proportion of the actual class included in the provided subregion.
To fulfill this, we employ the confidence scores predicted by the model as the source of judging the validity of the partial explanations.
The score for the target class is defined as follows:
\begin{equation}
\begin{split}
    Y^{k}_s &= \xi(\mathcal{L}_s, k) \\
    &=\textrm{exp} \left(\frac{\textrm{exp}(\mathcal{L}_s^{k,j})}{\sum_{i=1}^{N}\textrm{exp}(\mathcal{L}_s^{i,j})}\right) ,\forall j.
\end{split}
\end{equation}

Here, $j \in [0, n^2-1]$ indexes the logit, $\mathcal{L}_s$ is the output logit: $f(I_s)$ and $\xi(\cdot)$ returns the refined score with respect to the class $k$ by individually taking the exponential function to the softmax probability of $\mathcal{L}^j_s$, leading to expanding the gap of the partial confidence scores. $N$ represents the number of classes.
The final explanation is aggregated by applying the element-wise product among each partial explanation and the confidence scores.
The weighted partial explanations are scaled to the original patch size: $(n^2, 1, H', W')$ by the scaling factor $\beta$ to return to the original areas of the input image.
In summary, the partial explanation before the folding function is computed as follows:
\begin{equation}
    \mathbb{M}_s^k = \Psi(\mathbb{E}_s^k \odot Y_s^k, \beta)
\end{equation}

Then, the scaling function $\Psi(\cdot)$ is applied to interpolate the series of explanations with the size of $(H', W')$. Note that the map $\mathbb{E}_s^k$ is not normalized or clipped (eliminating the negative scores) to gather a more abundant range of explanations.
Furthermore, since the partial explanations share the overlapped portions from the single image, the frequency of the pixels in the saliency map is not accumulated evenly, destabilizing the resulting saliency map.
For this reason, we devise the duplication matrix $\Gamma$, which counts the duplication of each pixel. By utilizing this count, we average each pixel in the saliency map according to its frequency of occurrence.

Intuitively, the integration procedure is similar to the reverse of the unfolding process except for the average with duplication matrix. 
Including the frequency normalization with the duplication matrix $K$, we notate the integration procedure to generate the final saliency as follows:
\begin{equation}
    % M=\mathcal{V}(M_s, \rho)
    \mathbb{M}^k=\mathbb{V}(\mathbb{M}_s^k, \Gamma_s).
\end{equation}

The final saliency map has the same size as the input image, representing the relatively relevant scores in each pixel with respect to the output predictions of DNNs. The overall streams of our method are described in Algorithm.1.

\begin{algorithm}[t]
%%%%%%%%%%%%%%%%%%%%%%%%%%%%%%%%%%%%%%%%%%%%%%%%%%%%%%%%
%                     Algorithm                        %
%%%%%%%%%%%%%%%%%%%%%%%%%%%%%%%%%%%%%%%%%%%%%%%%%%%%%%%%
\caption{Saliency generation with UCAG}
\begin{algorithmic}[1]
\Require Image: $I$, Network: $f(\cdot)$, Explanation method: $\mathbb{E}$
Scaling factor: $\{\rho, \alpha, \beta\}$, Target class: $k$

% \Procedure $Unfold(I, \alpha, n)$ 
\Statex \textbf{Unfolding:} \Comment{Deploy a set of segments} 
\State $I_p = \mathbb{U}(I, \rho, n)$  \Comment{Striding pass} 
\State $I_s = \Psi(I_p, \alpha)$  \Comment{Scaling pass} 
\State $\mathcal{L}_s\leftarrow f(I_s)$  \Comment{Forward pass}
\Statex \textbf{Patch-wise Explanations:} \Comment{Generating saliencies}
\If{$\mathbb{E}$ is Gradient-based}
\State $\mathbb{E}_s\leftarrow \mathbb{E}_{grad}(I_s, f, k)$ \Comment{Utilize gradients}
\ElsIf{$\mathbb{E}$ is Propagation-based}
\State $\mathbb{E}_s\leftarrow \mathbb{E}_{att}(I_s, f, k)$ \Comment{Utilize propagations}
\EndIf
\Statex \textbf{Conquer:} \Comment{Integrating partial explanations}
\State $ Y^k_s \leftarrow \xi({L}_s, k) $ \Comment{Compute confidence scores} 
\State $ \mathbb{M}_s^k \leftarrow \Psi(\mathbb{E}_s^k \odot Y^k_s, \beta)$ \Comment{Aggregation and Scaling}
\State $ \mathbb{M}^k \leftarrow \mathbb{V}(\mathbb{M}_s^k, \Gamma_s)$ \Comment{Generating final explanation}

\label{alg:1}
\end{algorithmic}
\end{algorithm}

\section{Experimental Evaluation}
\subsection{Implementation Details}
We utilize the public accessible datasets: ImageNet-1k~\cite{deng2009imagenet}, Pascal VOC 2002~\cite{everingham2010pascal}, and MS COCO~\cite{lin2014microsoft}.
For the localization and segmentation assessments, we utilize the ImageNet segmentation dataset~\cite{guillaumin2014imagenet}. % 수정_ 4000장 있다는 내용 삭제
For fair comparisons in evaluations, we employ the online available models: VGG19~\cite{simonyan2014very}, DenseNet121~\cite{huang2017densely}, ResNet50~\cite{he2016deep}, and InceptionV3~\cite{szegedy2016rethinking}.
% We compare with i) gradient-based methods: GradCAM, XGradCAM, WGradCAM, GradCAM++, CAMERAS, NormGrad~\cite{rebuffi2019normgrad} and FullGrad~\cite{srinivas2019full}, ii) attribution-based methods: c*LRP, SGLRP, MWP, c*MWP, RAP, RSP, and AGF.

% Introducing evaluation metrics
% We utilized several metrics: i) insertion and deletion~\cite{petsiuk2018rise}, ii) (energy-based) pointing game~\cite{zhang2018top}, iii) mean Average Precision (mAP), iv) positive and negative map density~\cite{jalwana2021cameras}, to measure the degree of localization, causality, and density. A detailed comparison is described in each section.
We utilized several metrics: i) insertion and deletion games, ii) (energy-based) pointing game, iii) mean Average Precision (mAP), and iv) positive and negative map density to measure the degree of localization, causality, and density. A detailed comparison is described in each section.
We set the hyperparameter for the number of segments and rescaling factor as $(\alpha=2.6, n=6, \rho=0.555)$. Comparisons according to the hyper-parameter are provided in the supplementary materials.

\begin{figure}[t!]
\centering
\includegraphics[width=\linewidth]{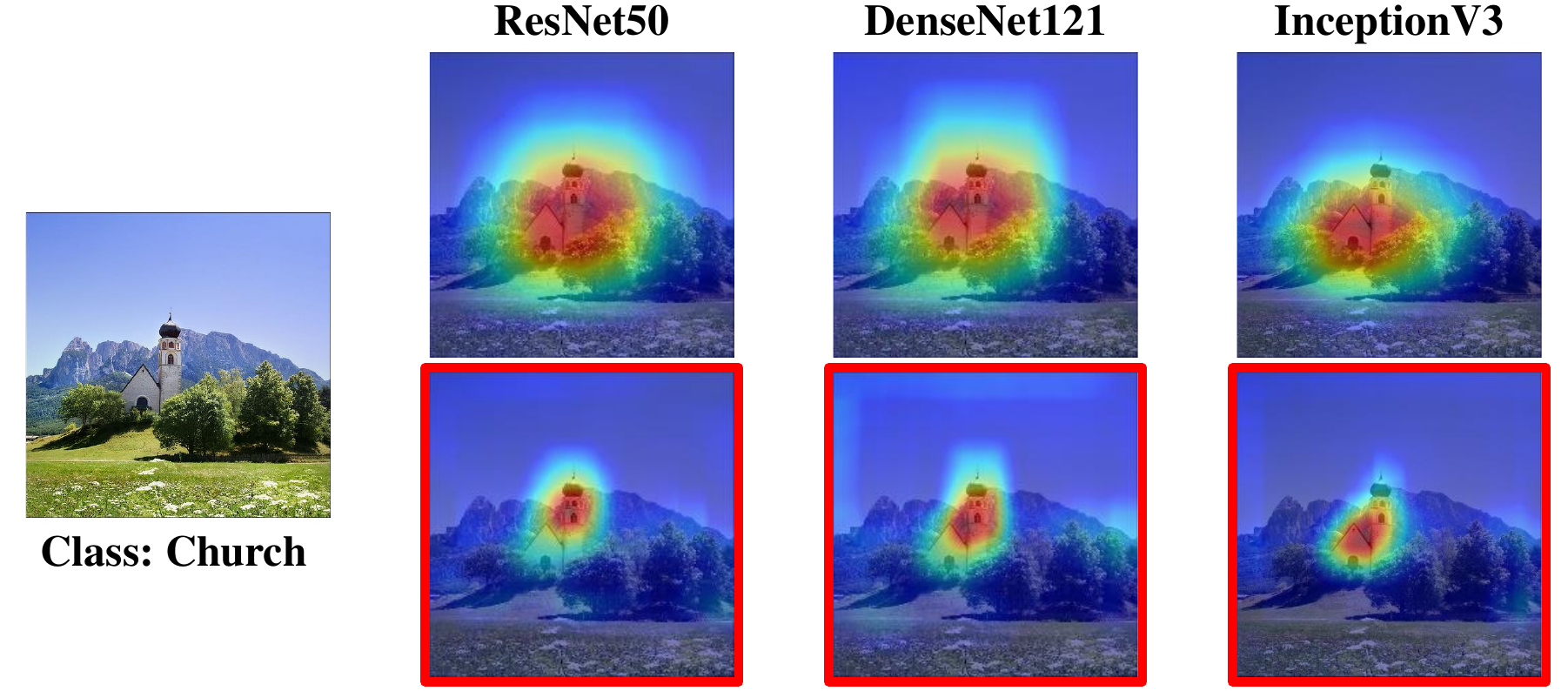}

\caption{Qualitative results between multiple networks including ResNet50, DenseNet121, and InceptionV3. The top and bottom rows represent visualized heatmaps generated by GradCAM with and without our approach, respectively. More results are contained in the supplementary material.}
% \caption{An overview of our approach. The CAM module takes the sub-regions divided from the single image and computes the outcomes independently. The computed outcomes are aggregated and attached into the single canvas. After the results are multiplied by the model's corresponding score, each result is aggregated and attached to a single canvas.}

\label{fig:aggregation}
\end{figure}
\subsection{Qualitative Results}
% - 시각적 결과 설명 -
In terms of explainability, it is crucial to provide intuitive and convincing visualizations from a human view. To demonstrate how our method improves the quality of generated explanations, we compare the results with the representative explanation methods in Figure ~\ref{fig:result1}. The first and second rows represent the results of the original and our methods, respectively. Red and blue colors denote the positive and negative relevance scores, respectively. While existing methods have the advantage of depicting the highly relevant regions to the output predictions, they still lack providing fine-grained descriptions with the limited granularity of representing areas of contribution.

In the manner of the post-hoc framework, UCAG has the strong advantages of fine-grained visualization, easily applicable and agnostic to model variations. In visualization, it improves the quality of existing maps by revising the incorrect negative or positive relevance scores in the (ir)relevant area without losing salient regions, resulting in increasing the density of the explanation map. Figure~\ref{fig:aggregation} shows the improvement when UCAG is applied to various models: ResNet, DenseNet, and InceptionNet. These visual enhancements also lead to improved quantitative performance. We report the additional visual explanations and evaluations on various methods and models, including GradCAM, FullGrad, GradCAM++, WGradCAM, RAP, RSP, AGF, SGLRP, and transformer interpretability ~\cite{chefer2021transformer}.

\subsection{Quantitative Results}

\begin{table}[!t]
\centering
\begin{centering}
% \begin{adjustbox}{width=.475\textwidth}
\begin{tabular}{l  c  c  c }
    \specialrule{2.0pt}{1pt}{1pt}
    % \cline{2-3}
    Methods & ResNet50 & DenseNet121 & InceptionV3 \\
    \hline
    GCAM & 11.3/53.9 & 10.6/48.1 & 10.0/52.8 \\ % 13.97/50.83
    GCAM++ & 11.6/52.7 & 10.9/47.2 & 10.1/52.0 \\ % 14.01/50.44
    WGCAM & 10.1/52.1 & 10.8/47.7 & 10.1/52.1 \\ % 15.23/48.86
    \cdashline{1-4}
    \textbf{GCAM*} & \textbf{8.6/53.4} & \textbf{8.9/49.0} & \textbf{8.59/53.4} \\ % 8.59/53.36
    \textbf{GCAM++*} & \textbf{8.7/52.7} &\textbf{9.2/48.0} & \textbf{8.86/51.9} \\ % 8.86/51.91
    \textbf{WGCAM*} & \textbf{9.1/52.8 }& \textbf{9.3/48.6} & \textbf{9.08/52.5} \\ 
    \specialrule{2.0pt}{1pt}{1pt}
\end{tabular}
\caption{The AUC scores regarding the deletion (lower is better)/insertion (higher is better) games on the ImageNet dataset. Mark * represents the performance of applying our methods.}
\label{table:ins_del}
% \end{adjustbox}
\end{centering}
\end{table}

% \begin{table}[!t]
% \centering
% \begin{centering}
% % \begin{adjustbox}{width=.475\textwidth}
% \begin{tabular}{l  c  c  c }
%     \specialrule{2.0pt}{1pt}{1pt}
%     % \cline{2-3}
%     Methods & *DeiT-Base & DeiT-Large & *ViT-Large \\
%     \hline
%     *Generic & 19.5/50.1 & - & 21.9/64.1 \\ % 13.97/50.83
%     *IBA & 15.6/54.1 & 17.4/\textbf{57.7} & 20.6/66.7    \\ % 14.01/50.44
%     Ours & 15.1/54.8 & 17.2/57.2 & 19.8/66.9  \\ % 15.23/48.86
%     \textbf{Ours + SPL} & \textbf{14.4/55.2} &\textbf{ 16.8}/57.4 & \textbf{18.2/68.1}  \\ % 15.23/48.86
%     \specialrule{2.0pt}{1pt}{1pt}
% \end{tabular}
% \caption{The AUC scores regarding the deletion (lower is better)/insertion (higher is better) games on the ImageNet dataset. Mark * represents the performance of applying our methods.}
% \label{table:ins_del}
% % \end{adjustbox}
% \end{centering}
% \end{table}

\begin{table}[t!]
\centering
\begin{centering}
\renewcommand{\arraystretch}{1.0} 
% \begin{adjustbox}{width=.475\textwidth}
\begin{tabular}{p{1.4cm}  p{1.25 cm}  p{1.25 cm}  p{1.25 cm}  p{1.25 cm}}
     \specialrule{2.0pt}{1pt}{1pt} % (all/diff)
    &  \multicolumn{2}{c}{\underline{VOC07 Test}} & \multicolumn{2}{c}{\underline{COCO14 Val}} \\
    % \cline{2-3}
    Methods & \begin{tabular}[c]{c@{}}VGG16\end{tabular} & ResNet50 & \begin{tabular}[c]{c@{}}VGG16\end{tabular} & ResNet50 \\
    \hline
    Center & 69.6/42.4 & 69.6/42.4 & 27.8/19.5 & 27.8/19.5 \\
    Gradient & 76.3/56.9 & 72.3/56.8 & 37.7/31.4 & 35.0/29.4 \\
    DeConv & 67.5/44.2 & 68.6/44.7 & 30.7/23.0 & 30.0/21.9 \\
    Guid & 75.9/53.0 & 77.2/59.4 & 39.1/31.4 & 42.1/35.3 \\
    MWP & 77.1/56.6 & 84.4/70.8 & 39.8/32.8 & 49.6/43.9 \\
    cMWP & 79.9/66.5 & 90.7/82.1 & 49.7/44.3 & 58.5/53.6 \\
    RISE & 86.9/75.1 & 86.4/78.8 & 50.8/45.3 & 54.7/50.0 \\
    GradCAM & 86.6/74.0 & 90.4/82.3 & 54.2/49.0 & 57.3/52.3 \\
    Extremal & 88.0/76.1 & 88.9/78.7 & 51.5/45.9 & 56.5/51.5 \\
    NormGrad & 81.9/64.8 & 84.6/72.2 & - & - \\
    CAMERAS & 86.2/76.2 & 94.2/88.8 & 55.4/50.7 & 69.6/66.4 \\
    \textbf{Ours} & \textbf{91.1/82.8} & \textbf{94.2/89.4} & \textbf{61.8/57.6} & \textbf{71.0/67.6} \\
    \specialrule{2.0pt}{1pt}{1pt}
\end{tabular}

% \end{adjustbox}
\end{centering}
\caption{The performance of the pointing game among various methods. Our method (applied to the GradCAM) represents a sizeable increment in performance compared to the other method.}
\label{table:pg}
\end{table}
\subsubsection{Causality assessment}

The deletion and insertion game measures whether the explanation map correctly highlights areas contributing to the decision-making process. Deletion measures the model's response by repeatedly deleting the high-priority pixels with respect to the output. When deleted pixels contribute significantly to predictions, model confidence and deletion scores decrease dramatically. In contrast, the insertion game measures how the confidence score would be increased.
We set the baseline image blurred with kernel size: $51$ and $\sigma$: $50$ for the insertion game and images with zero pixels for the deletion game.

Table \ref{table:ins_del} displays the scores of class activation maps with/without UCAG. The results show that the saliency map passed through the UCAG yields a decreased deletion score while the insertion score is maintained or increased. Specifically, the spatial scrutinization in UCAG successfully induces the outcomes to be more precise while preserving saliency.

\begin{table*}[h!]
\centering
\begin{centering}
\begin{tabular}{lccccc}
\specialrule{2.0pt}{1pt}{1pt}
Model                   & Dataset                                                        & GCAM                  & GCAM++                & XGCAM                 & WGCAM                  \\ \hline
\multirow{2}{*}{VGG}    & \begin{tabular}[c]{@{}c@{}}VOC07 Test\end{tabular} & 56.6(40.1)/\textbf{59.7(43.6)} & 53.7(36.0)/\textbf{61.4(47.6)} & 55.6(39.6)/\textbf{59.3(43.2)} & 55.6(37.4)/\textbf{56.5(38.5)} \\
                        & \begin{tabular}[c]{@{}c@{}}COCO14 Val\end{tabular} & 25.5(19.6)/\textbf{29.9(23.8)} & 22.4(16.5)/\textbf{29.2(23.3)} & 25.0(19.2)/\textbf{28.6(22.6)} & 24.4(18.1)/\textbf{26.8(20.3)} \\ \hline
\multirow{2}{*}{ResNet} & \begin{tabular}[c]{@{}c@{}}VOC07 Test\end{tabular} & 61.3(45.3)/\textbf{74.0(66.8)} & 52.9(31.8)/\textbf{63.9(47.6)} & 61.3(45.3)/\textbf{74.0(66.8)} & 55.9(36.8)/\textbf{64.2(44.9)} \\
                        & \begin{tabular}[c]{@{}c@{}}COCO14 Val\end{tabular} & 27.5(21.1)/\textbf{42.7(37.4)} & 21.9(15.2)/\textbf{31.4(25.0)} & 27.5(21.1)/\textbf{42.7(37.4)} & 23.1(16.5)/\textbf{35.4(29.6)} \\ 

\specialrule{2.0pt}{1pt}{1pt}
\end{tabular}
% \end{adjustbox}
\caption{The performance of the energy-based pointing game represented as original / ours. $(-)$ denotes the result in the difficult set of each dataset.}
\label{table:epg}
\end{centering}
\end{table*}

\begin{table*}[!tbp]
\centering
\resizebox{1\textwidth}{!}{
\begin{tabular}{ccccccccc}
\specialrule{2.0pt}{1pt}{1pt}
    Model      &     Metric   & FullGrad & LRP & c*LRP & SGLRP & RAP & RSP  & AGF \\
\midrule
% \multirow{2}{*}{VGG}    & mAP    
% & 0.7479/0.7744 &  &  &  &  & \\
%           & Out/In 
% & 0.7981/0.8206 &  &  &  &  &  \\
% pixAcc: 0.7007, mAP: 0.6599:
\multirow{2}{*}{VGG}    & mAP    
& \textbf{0.789}/0.783 & \textbf{0.700}/0.682 & 0.547/\textbf{0.559} & 0.551/\textbf{0.570} & 0.735/\textbf{0.743} & 0.743/\textbf{0.752} & 0.740/\textbf{0.752} \\
          & Pixel Acc. 
& \textbf{0.742}/0.737 & \textbf{0.750}/0.730 & 0.534/\textbf{0.556} & 0.543/\textbf{0.587} & 0.786/\textbf{0.789} & 0.793/\textbf{0.796} & 0.779/\textbf{0.794} \\
\midrule
% 0.7369, mAP: 0.6892
\multirow{2}{*}{ResNet} & mAP
& 0.748/\textbf{0.774} & 0.717/\textbf{0.722} & 0.671/\textbf{0.694} & 0.588/\textbf{0.617} & 0.711/\textbf{0.761} & 0.719/\textbf{0.766} &  0.758/\textbf{0.783}\\
          & Pixel Acc.
& 0.798/\textbf{0.821} & 0.770/\textbf{0.774} & 0.716/\textbf{0.736} & 0.608/\textbf{0.648} & 0.763/\textbf{0.809} & 0.775/\textbf{0.813} & 0.804/\textbf{0.826} \\
\specialrule{2.0pt}{1pt}{1pt}
\end{tabular}
}
\caption{Performance of mAP and outside--inside ratio over ImageNet segmentation dataset (Original / Ours). Propagation-based methods that have the class-discriminative ability are compared. A Low Out/In ratio denotes attributions are intensively distributed in the semantic mask.} \label{table:tab2}
\label{table:lrp_seg}
\end{table*}

\subsubsection{Localization assessment}
Evaluating localization performance is employed to assess the quality of the explanation, assuming that the highlighted pixels contain the main object of the input image. 
\cite{zhang2018top} introduced the Pointing Game, widely used in many related researches, that computes whether the maximum point of the saliency map is contained in the label. We report our results (GradCAM with UCAG) in Table \ref{table:pg}, and it outperforms the state-of-art explanatory approaches. 

ScoreCAM~\cite{wang2020score} raised the issue that the pointing game does not provide a detailed measure of localization performance since it only considers the maximum saliency point inside the explanation. As an alternative, ScoreCAM introduces an energy-based pointing game to evaluate the quality with a more detailed measurement. 
The energy-based pointing game is based on the summation of the regions where saliency overlaps the inside label.
We also report the comparisons of the energy-based pointing game in Table \ref{table:epg} with the results of various CAM approaches. 
The increased scores indicate that UCAG consistently enhances the ability of various CAM approaches regardless of the type of network and method.

Evaluating the weakly supervised segmentation is closely related to interpretability in terms of finding the salient objects in the input image. In recent works~\cite{nam2020relative, nam2021interpreting, gur2021visualization}, mean average precision (mAP) is utilized to assess the quality of interpretability. We follow the experiential settings with the ImageNet segmentation dataset~\cite{guillaumin2014imagenet}. Table~\ref{table:lrp_seg} shows the comparisons of existing methods and ours. In most cases, our method improves the quality of interpretations in both VGG and ResNet models.

\begin{table}[t!]
\centering
% \begin{adjustbox}{width=0.5\textwidth}
\begin{tabular}{p{3.3cm}ccc}
\specialrule{2.0pt}{1pt}{1pt}
Model   & GCAM   & CAMERAS & Ours \\ \hline
\multicolumn{4}{l}{Positive map density $(\mathbb{D}_{map}^{+} \uparrow{})$}   \\ \hline
ResNet50         & 2.33            & 3.20             & \textbf{3.89} \\
DenseNet         & 2.35            & 3.23             & \textbf{3.45} \\
Inceptionv3      & 2.18            & 3.15             & \textbf{3.83} \\ \hline
\multicolumn{4}{l}{Negative map density $(\mathbb{D}_{map}^{-} \downarrow{})$} \\ \hline
ResNet50         & 0.86            & 0.81    & \textbf{0.81} \\
DenseNet         & 0.94            & 0.83             & \textbf{0.82} \\
Inceptionv3      & 1.04            & 0.93             & \textbf{0.85} \\ 
\specialrule{2.0pt}{1pt}{1pt}
\end{tabular}
\caption{Evaluated results of positive (higher is better) and negative (lower is better) map density.}
% \end{adjustbox}
% \end{centering}
\label{table:pmd}
\end{table}

\subsubsection{Density assessment}
Several previous works~\cite{jalwana2021cameras, poppi2021revisiting} pointed out that only reporting the ratio of (de)increase in confidence score does not fully reflect the interpretability of explaining methods.
As an alternative, CAMERAS~\cite{jalwana2021cameras} introduces the positive (negative) map density, which simultaneously evaluates the variations of confidence scores and saliency density during the perturbation.
Metrics for positive (negative) map density are defined as $\mathbb{D}_{map}^+ = f(I{\odot}\mathbb{M}^k)/\sum_i{\sum_j{\mathbb{M}^k_{(i,j)}}}\times(h \times w)$ and $\mathbb{D}_{map}^- = f(I{\odot}1-\mathbb{M}^k)/{\sum_i{\sum_j{(1-\mathbb{M}^k_{(i,j)})}}}\times(h \times w)$. 
% As the metric provides the density of saliency and increment of prediction score, it is possible to measure the quality of the explanation. % 수정_ 이 문장 삭제
Therefore, it is possible to measure the quality of the explanation in a more precise way.

The results in Table \ref{table:pmd} show that UCAG enhances the density of saliency maps more than the existing multi-resolution fusion approach. This implies that our approach has the advantage of strong objectness with concentrated scores in the saliency regions. 

\subsection{Ablation and Discussion}
As an ablation study, the gap of effectiveness between upscaling and spatial scrutinization is described in Figure~\ref{fig:ins_del}. 
% UCAG induces the explanation method to decrease the deletion scores while amplifying the insertion score. 
% In other words, spatial scrutinization provides a more precise explanation with the tendency to preserve the areas contributing to the decision-making process.
UCAG provides a more precise explanation with the tendency to preserve the areas contributing to the decision-making process. % 수정_ In other words 문장 지우고 UCAG를 이문장에 직접 삽입

Here, we note that the computational cost is inevitably increased due to the manner of post-hoc framework. However, it is more efficient to find various spatial saliency with the scrutinization rather than the cost of calculating the exponentially increasing image resolution. In addition, we implement the overall frameworks batch-wisely, leading to enhancing computational efficiency.

% - resize vs ours -> density, causality는 성능으로, localization만 정성첨가

% However, from the measurement, we found that the desirable hyper-parameters differ based on the network type. For example, for the densenet121, the hyper-parameter combination of $(124, 124, 6, 2.6)$  for $(p_h, p_w, n, \alpha)$. However, for the inceptionv3, the combination $(124, 124, 6, 2.6)$ generates better results for several measurements, such as deletion, insertion, and pointing games. Therefore, to reduce the bias of hyper-parameters between networks, it is necessary to reveal the relationship between network properties and approaches.

% \begin{figure}[!tbp]
% % \centering
% \includegraphics[width=0.475\textwidth]{./figs/figure-ps.png}
% % \includegraphics[width=1.0\textwidth]{./figs/methods-overview.png}
% % \includegraphics[width=1.0\textwidth]{./figs/methods.png}

% \caption{Visualized explanation maps contain the visualized explanation maps computed by divergent patch sizes.}
% % \caption{An overview of our approach. The CAM module takes the sub-regions divided from the single image and computes the outcomes independently. The computed outcomes are aggregated and attached into the single canvas. After the results are multiplied by the model's corresponding score, each result is aggregated and attached to a single canvas.}

% \label{fig:patch_size}
% \end{figure}

\section{Conclusion}
In this paper, we propose a post-hoc framework: UCAG, to enhance the interpretability of network decisions by guiding the explaining methods to carefully investigate the saliencies with spatial scrutinization and aggregation.
We demonstrate the validity of UCAG from qualitative and quantitative results. With the advantage of being agnostic to the model and method, the increments of causality, localization, and density demonstrate the effectiveness of UCAG for producing a better visual representation ability of network decisions.

\section{Acknowledgements}
This work was supported by Institute of Information \& communications Technology Planning \& Evaluation (IITP) grant funded by the Korea government(MSIT) (No.2022-0-00984, Development of Artificial Intelligence Technology for Personalized Plug-and-Play Explanation and Verification of Explanation \& No. 2019-0-00079, Artificial Intelligence Graduate School Program, Korea University)

\bibliography{aaai23}

\begin{thebibliography}{34}
\providecommand{\natexlab}[1]{#1}

\bibitem[{Bach et~al.(2015)Bach, Binder, Montavon, Klauschen, M{\"u}ller, and Samek}]{bach2015pixel}
Bach, S.; Binder, A.; Montavon, G.; Klauschen, F.; M{\"u}ller, K.-R.; and Samek, W. 2015.
\newblock On pixel-wise explanations for non-linear classifier decisions by layer-wise relevance propagation.
\newblock \emph{PloS one}, 10(7): e0130140.

\bibitem[{Chattopadhay et~al.(2018)Chattopadhay, Sarkar, Howlader, and Balasubramanian}]{chattopadhay2018grad}
Chattopadhay, A.; Sarkar, A.; Howlader, P.; and Balasubramanian, V.~N. 2018.
\newblock Grad-cam++: Generalized gradient-based visual explanations for deep convolutional networks.
\newblock In \emph{2018 IEEE winter conference on applications of computer vision (WACV)}, 839--847. IEEE.

\bibitem[{Chefer, Gur, and Wolf(2021)}]{chefer2021transformer}
Chefer, H.; Gur, S.; and Wolf, L. 2021.
\newblock Transformer interpretability beyond attention visualization.
\newblock In \emph{Proceedings of the IEEE/CVF Conference on Computer Vision and Pattern Recognition}, 782--791.

\bibitem[{Deng et~al.(2009)Deng, Dong, Socher, Li, Li, and Fei-Fei}]{deng2009imagenet}
Deng, J.; Dong, W.; Socher, R.; Li, L.-J.; Li, K.; and Fei-Fei, L. 2009.
\newblock Imagenet: A large-scale hierarchical image database.
\newblock In \emph{2009 IEEE conference on computer vision and pattern recognition}, 248--255. Ieee.

\bibitem[{Everingham et~al.(2010)Everingham, Van~Gool, Williams, Winn, and Zisserman}]{everingham2010pascal}
Everingham, M.; Van~Gool, L.; Williams, C.~K.; Winn, J.; and Zisserman, A. 2010.
\newblock The pascal visual object classes (voc) challenge.
\newblock \emph{International journal of computer vision}, 88(2): 303--338.

\bibitem[{Fong, Patrick, and Vedaldi(2019)}]{fong2019understanding}
Fong, R.; Patrick, M.; and Vedaldi, A. 2019.
\newblock Understanding deep networks via extremal perturbations and smooth masks.
\newblock In \emph{Proceedings of the IEEE/CVF international conference on computer vision}, 2950--2958.

\bibitem[{Fu et~al.(2020)Fu, Hu, Dong, Guo, Gao, and Li}]{fu2020axiom}
Fu, R.; Hu, Q.; Dong, X.; Guo, Y.; Gao, Y.; and Li, B. 2020.
\newblock Axiom-based grad-cam: Towards accurate visualization and explanation of cnns.
\newblock \emph{arXiv preprint arXiv:2008.02312}.

\bibitem[{Gu, Yang, and Tresp(2018)}]{gu2018understanding}
Gu, J.; Yang, Y.; and Tresp, V. 2018.
\newblock Understanding individual decisions of cnns via contrastive backpropagation.
\newblock In \emph{Asian Conference on Computer Vision}, 119--134. Springer.

\bibitem[{Guillaumin, K{\"u}ttel, and Ferrari(2014)}]{guillaumin2014imagenet}
Guillaumin, M.; K{\"u}ttel, D.; and Ferrari, V. 2014.
\newblock Imagenet auto-annotation with segmentation propagation.
\newblock \emph{International Journal of Computer Vision}, 110(3): 328--348.

\bibitem[{Gur, Ali, and Wolf(2021)}]{gur2021visualization}
Gur, S.; Ali, A.; and Wolf, L. 2021.
\newblock Visualization of supervised and self-supervised neural networks via attribution guided factorization.
\newblock In \emph{Proceedings of the AAAI Conference on Artificial Intelligence}, volume~35, 11545--11554.

\bibitem[{He et~al.(2016)He, Zhang, Ren, and Sun}]{he2016deep}
He, K.; Zhang, X.; Ren, S.; and Sun, J. 2016.
\newblock Deep residual learning for image recognition.
\newblock In \emph{Proceedings of the IEEE conference on computer vision and pattern recognition}, 770--778.

\bibitem[{Huang et~al.(2017)Huang, Liu, Van Der~Maaten, and Weinberger}]{huang2017densely}
Huang, G.; Liu, Z.; Van Der~Maaten, L.; and Weinberger, K.~Q. 2017.
\newblock Densely connected convolutional networks.
\newblock In \emph{Proceedings of the IEEE conference on computer vision and pattern recognition}, 4700--4708.

\bibitem[{Iwana, Kuroki, and Uchida(2019)}]{iwana2019explaining}
Iwana, B.~K.; Kuroki, R.; and Uchida, S. 2019.
\newblock Explaining convolutional neural networks using softmax gradient layer-wise relevance propagation.
\newblock In \emph{2019 IEEE/CVF International Conference on Computer Vision Workshop (ICCVW)}, 4176--4185. IEEE.

\bibitem[{Jalwana et~al.(2021)Jalwana, Akhtar, Bennamoun, and Mian}]{jalwana2021cameras}
Jalwana, M.~A.; Akhtar, N.; Bennamoun, M.; and Mian, A. 2021.
\newblock CAMERAS: Enhanced resolution and sanity preserving class activation mapping for image saliency.
\newblock In \emph{Proceedings of the IEEE/CVF Conference on Computer Vision and Pattern Recognition}, 16327--16336.

\bibitem[{Jiang et~al.(2021)Jiang, Zhang, Hou, Cheng, and Wei}]{jiang2021layercam}
Jiang, P.-T.; Zhang, C.-B.; Hou, Q.; Cheng, M.-M.; and Wei, Y. 2021.
\newblock Layercam: Exploring hierarchical class activation maps for localization.
\newblock \emph{IEEE Transactions on Image Processing}, 30: 5875--5888.

\bibitem[{Jung and Oh(2021)}]{jung2021towards}
Jung, H.; and Oh, Y. 2021.
\newblock Towards better explanations of class activation mapping.
\newblock In \emph{Proceedings of the IEEE/CVF International Conference on Computer Vision}, 1336--1344.

\bibitem[{Lin et~al.(2014)Lin, Maire, Belongie, Hays, Perona, Ramanan, Doll{\'a}r, and Zitnick}]{lin2014microsoft}
Lin, T.-Y.; Maire, M.; Belongie, S.; Hays, J.; Perona, P.; Ramanan, D.; Doll{\'a}r, P.; and Zitnick, C.~L. 2014.
\newblock Microsoft coco: Common objects in context.
\newblock In \emph{European conference on computer vision}, 740--755. Springer.

\bibitem[{Montavon et~al.(2017)Montavon, Lapuschkin, Binder, Samek, and M{\"u}ller}]{montavon2017explaining}
Montavon, G.; Lapuschkin, S.; Binder, A.; Samek, W.; and M{\"u}ller, K.-R. 2017.
\newblock Explaining nonlinear classification decisions with deep taylor decomposition.
\newblock \emph{Pattern recognition}, 65: 211--222.

\bibitem[{Nam, Choi, and Lee(2021)}]{nam2021interpreting}
Nam, W.-J.; Choi, J.; and Lee, S.-W. 2021.
\newblock Interpreting deep neural networks with relative sectional propagation by analyzing comparative gradients and hostile activations.
\newblock In \emph{Proceedings of the AAAI Conference on Artificial Intelligence}, volume~35, 11604--11612.

\bibitem[{Nam et~al.(2020)Nam, Gur, Choi, Wolf, and Lee}]{nam2020relative}
Nam, W.-J.; Gur, S.; Choi, J.; Wolf, L.; and Lee, S.-W. 2020.
\newblock Relative attributing propagation: Interpreting the comparative contributions of individual units in deep neural networks.
\newblock In \emph{Proceedings of the AAAI Conference on Artificial Intelligence}, volume~34, 2501--2508.

\bibitem[{Petsiuk, Das, and Saenko(2018)}]{petsiuk2018rise}
Petsiuk, V.; Das, A.; and Saenko, K. 2018.
\newblock Rise: Randomized input sampling for explanation of black-box models.
\newblock \emph{arXiv preprint arXiv:1806.07421}.

\bibitem[{Poppi et~al.(2021)Poppi, Cornia, Baraldi, and Cucchiara}]{poppi2021revisiting}
Poppi, S.; Cornia, M.; Baraldi, L.; and Cucchiara, R. 2021.
\newblock Revisiting the evaluation of class activation mapping for explainability: A novel metric and experimental analysis.
\newblock In \emph{Proceedings of the IEEE/CVF Conference on Computer Vision and Pattern Recognition}, 2299--2304.

\bibitem[{Rebuffi et~al.(2019)Rebuffi, Fong, Ji, Bilen, and Vedaldi}]{rebuffi2019normgrad}
Rebuffi, S.-A.; Fong, R.; Ji, X.; Bilen, H.; and Vedaldi, A. 2019.
\newblock NormGrad: Finding the pixels that matter for training.
\newblock \emph{arXiv preprint arXiv:1910.08823}.

\bibitem[{Rebuffi et~al.(2020)Rebuffi, Fong, Ji, and Vedaldi}]{rebuffi2020there}
Rebuffi, S.-A.; Fong, R.; Ji, X.; and Vedaldi, A. 2020.
\newblock There and back again: Revisiting backpropagation saliency methods.
\newblock In \emph{Proceedings of the IEEE/CVF Conference on Computer Vision and Pattern Recognition}, 8839--8848.

\bibitem[{Selvaraju et~al.(2017)Selvaraju, Cogswell, Das, Vedantam, Parikh, and Batra}]{selvaraju2017grad}
Selvaraju, R.~R.; Cogswell, M.; Das, A.; Vedantam, R.; Parikh, D.; and Batra, D. 2017.
\newblock Grad-cam: Visual explanations from deep networks via gradient-based localization.
\newblock In \emph{Proceedings of the IEEE international conference on computer vision}, 618--626.

\bibitem[{Simonyan, Vedaldi, and Zisserman(2013)}]{simonyan2013deep}
Simonyan, K.; Vedaldi, A.; and Zisserman, A. 2013.
\newblock Deep inside convolutional networks: Visualising image classification models and saliency maps.
\newblock \emph{arXiv preprint arXiv:1312.6034}.

\bibitem[{Simonyan and Zisserman(2014)}]{simonyan2014very}
Simonyan, K.; and Zisserman, A. 2014.
\newblock Very deep convolutional networks for large-scale image recognition.
\newblock \emph{arXiv preprint arXiv:1409.1556}.

\bibitem[{Springenberg et~al.(2014)Springenberg, Dosovitskiy, Brox, and Riedmiller}]{springenberg2014striving}
Springenberg, J.~T.; Dosovitskiy, A.; Brox, T.; and Riedmiller, M. 2014.
\newblock Striving for simplicity: The all convolutional net.
\newblock \emph{arXiv preprint arXiv:1412.6806}.

\bibitem[{Srinivas and Fleuret(2019)}]{srinivas2019full}
Srinivas, S.; and Fleuret, F. 2019.
\newblock Full-gradient representation for neural network visualization.
\newblock \emph{Advances in neural information processing systems}, 32.

\bibitem[{Szegedy et~al.(2016)Szegedy, Vanhoucke, Ioffe, Shlens, and Wojna}]{szegedy2016rethinking}
Szegedy, C.; Vanhoucke, V.; Ioffe, S.; Shlens, J.; and Wojna, Z. 2016.
\newblock Rethinking the inception architecture for computer vision.
\newblock In \emph{Proceedings of the IEEE conference on computer vision and pattern recognition}, 2818--2826.

\bibitem[{Wang et~al.(2020)Wang, Wang, Du, Yang, Zhang, Ding, Mardziel, and Hu}]{wang2020score}
Wang, H.; Wang, Z.; Du, M.; Yang, F.; Zhang, Z.; Ding, S.; Mardziel, P.; and Hu, X. 2020.
\newblock Score-CAM: Score-weighted visual explanations for convolutional neural networks.
\newblock In \emph{Proceedings of the IEEE/CVF conference on computer vision and pattern recognition workshops}, 24--25.

\bibitem[{Zeiler and Fergus(2014)}]{zeiler2014visualizing}
Zeiler, M.~D.; and Fergus, R. 2014.
\newblock Visualizing and understanding convolutional networks.
\newblock In \emph{European conference on computer vision}, 818--833. Springer.

\bibitem[{Zhang et~al.(2018)Zhang, Bargal, Lin, Brandt, Shen, and Sclaroff}]{zhang2018top}
Zhang, J.; Bargal, S.~A.; Lin, Z.; Brandt, J.; Shen, X.; and Sclaroff, S. 2018.
\newblock Top-down neural attention by excitation backprop.
\newblock \emph{International Journal of Computer Vision}, 126(10): 1084--1102.

\bibitem[{Zhang, Rao, and Yang(2021)}]{zhang2021novel}
Zhang, Q.; Rao, L.; and Yang, Y. 2021.
\newblock A novel visual interpretability for deep neural networks by optimizing activation maps with perturbation.
\newblock In \emph{Proceedings of the AAAI Conference on Artificial Intelligence}, volume~35, 3377--3384.

\end{thebibliography}
\nocite{*}
\end{document}